\documentclass[sigconf]{acmart}
\usepackage{microtype}
\usepackage{graphicx}
\usepackage{booktabs} %
\usepackage{multirow}
\usepackage{pifont}
\usepackage{float}
\usepackage{natbib}
\usepackage{adjustbox}
\usepackage{enumitem}
\setlist[itemize]{leftmargin=*}

\usepackage{hyperref}

\usepackage{bbm}

\usepackage{amsmath}
\usepackage{mathtools}
\usepackage{amsthm}
\usepackage{multirow}

\usepackage[capitalize]{cleveref}
\usepackage{subcaption}
\theoremstyle{plain}

\theoremstyle{definition}

\theoremstyle{remark}

\usepackage[textsize=tiny]{todonotes}

\usepackage{soul}
\usepackage{xspace}
\definecolor{navy}{rgb}{0.1, 0.1, 0.8}
\definecolor[named]{gray}{rgb}{0.4, 0.4, 0.4}
\definecolor[named]{olive}{rgb}{0.1, 0.5, 0.1}
\definecolor[named]{ruby}{rgb}{0.8, 0.1, 0.3}
\definecolor{darkpastelgreen}{rgb}{0.01, 0.75, 0.24}
\definecolor{celestialblue}{rgb}{0.29, 0.59, 0.82}
\definecolor{coral}{rgb}{1.0, 0.5, 0.31}
\definecolor{Goldenrod}{rgb}{0.8,0.8,0}

\newcommand{\BLTPP}{\textsc{Language-TPP}}

\title{Byte-token Enhanced Language Models for Temporal Point Processes Analysis}

\author{Quyu Kong}
\email{kongquyu@gmail.com}
\affiliation{%
  \institution{Independent Researcher}
  \city{Hangzhou}
  \country{China}}

\author{Yixuan Zhang}
\email{zh1xuan@hotmail.com}
\affiliation{%
  \institution{Southeast University}
  \city{Nanjing}
  \country{China}}

\author{Yang Liu}
\email{lyng_95@zju.edu.cn}
\affiliation{%
  \institution{Independent Researcher}
  \city{Hangzhou}
  \country{China}}

\author{Panrong Tong}
\email{tongpanrong@hotmail.com}
\affiliation{%
  \institution{Independent Researcher}
  \city{Hangzhou}
  \country{China}}

\author{Enqi Liu}
\email{leq627@126.com}
\affiliation{%
  \institution{Independent Researcher}
  \city{Hangzhou}
  \country{China}}

\author{Feng Zhou}
\authornote{Corresponding author.}
\email{feng.zhou@ruc.edu.cn}
\affiliation{%
  \institution{Center for Applied Statistics and School of Statistics, Renmin University of China}
  \city{Beijing}
  \country{China}}

\copyrightyear{2026}
\acmYear{2026}
\setcopyright{cc}
\setcctype{by}
\acmConference[WWW '26]{Proceedings of the ACM Web Conference 2026}{April 13--17, 2026}{Dubai, United Arab Emirates}
\acmBooktitle{Proceedings of the ACM Web Conference 2026 (WWW '26), April 13--17, 2026, Dubai, United Arab Emirates}
\acmPrice{}
\acmDOI{10.1145/3774904.3792197}
\acmISBN{979-8-4007-2307-0/2026/04}


\begin{document}

\begin{abstract}
Temporal Point Processes (TPPs) have been widely used for modeling event sequences on the Web, such as user reviews, social media posts, and online transactions. However, traditional TPP models often struggle to effectively incorporate the rich textual descriptions that accompany these events, while Large Language Models (LLMs), despite their remarkable text processing capabilities, lack mechanisms for handling the temporal dynamics inherent in Web-based event sequences. To bridge this gap, we introduce \textbf{\BLTPP{}}, a unified framework that seamlessly integrates TPPs with LLMs for enhanced Web event sequence modeling. Our key innovation is a novel temporal encoding mechanism that converts continuous time intervals into specialized byte-tokens, enabling direct integration with standard language model architectures for TPP modeling without requiring TPP-specific modifications. This approach allows \BLTPP{} to achieve state-of-the-art performance across multiple TPP benchmarks, including event time prediction and type prediction, on real-world Web datasets spanning e-commerce reviews, social media and online Q\&A platforms. More importantly, we demonstrate that our unified framework unlocks new capabilities for TPP research: incorporating temporal information improves the quality of generated event descriptions, as evidenced by enhanced ROUGE-L scores, and better aligned sentiment distributions. Through comprehensive experiments, including qualitative analysis of learned distributions and scalability evaluations on long sequences, we show that \BLTPP{} effectively captures both temporal dynamics and textual patterns in Web user behavior, with important implications for content generation, user behavior understanding, and Web platform applications. Code is available at \url{https://github.com/qykong/Language-TPP}.

\end{abstract}
\maketitle
\section{Introduction}

Temporal Point Processes (TPPs) provide a statistical framework for modeling sequences of events occurring in continuous time on the Web. Traditional TPP models have been proven effective in capturing temporal dynamics and event types across diverse Web applications, including social media information diffusion~\citep{farajtabar2015coevolve,Mishra2016FeaturePrediction}, online user behavior modeling~\citep{kong2023interval}, and e-commerce platforms~\citep{xue2023easytpp}. However, these models have primarily focused on temporal and categorical aspects while the rich multi-modal information inherent in real-world online events has been under-explored.

Many Web events are accompanied by textual information that provides essential context beyond timestamps and event types. For example, product reviews on e-commerce platforms like Amazon~\citep{ni2019justifying} contain detailed user opinions, posts on social media platforms like Twitter include rich textual content~\citep{kong2020describing}, and questions on community Q\&A sites like Stack Overflow provide technical descriptions. The ability to model and, more importantly, generate such multi-modal event descriptions represents a significant opportunity for understanding Web user behavior and improving Web content generation—aspects that have been under-explored in previous TPP research.

Meanwhile, Large Language Models (LLMs) have achieved success in understanding and generating textual modality across numerous domains~\citep{jin2023time,m2024augmenting,xin2024bioinformatics}. This advancement allows to develop a unified framework between TPPs and LLMs that jointly models both temporal dynamics and textual information in event sequences. 
In this work, we aim to address the following two open questions. 

\begin{figure}[t]
    \centering
    \includegraphics[width=0.4\textwidth,page=1]{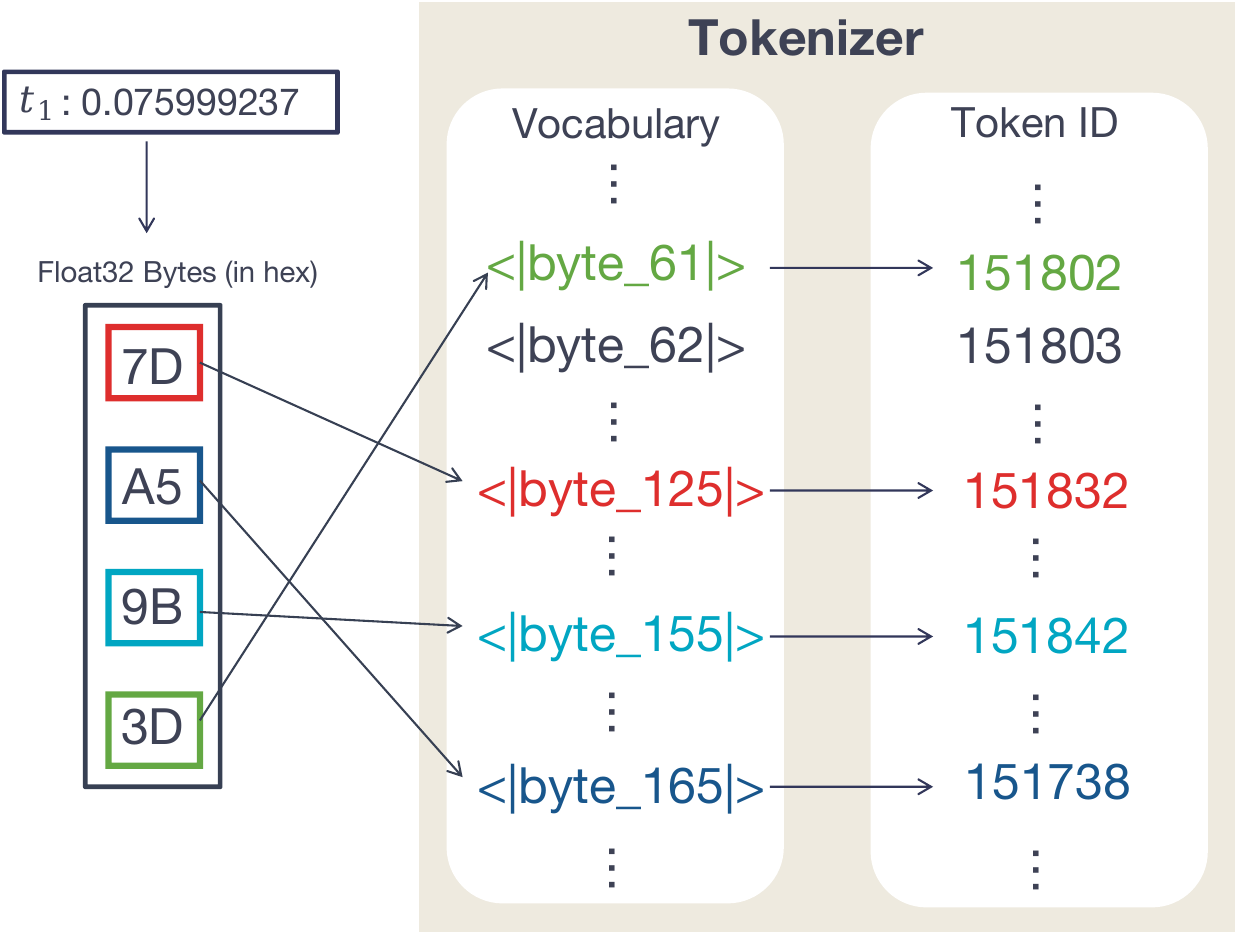}%
    \caption{Temporal tokenization procedure in \BLTPP{}. The diagram illustrates how event times are converted into temporal byte tokens for processing by the model.}
    \label{fig:temporal_tokenization}
    \vspace{-0.5cm}
\end{figure}

While prior works have successfully incorporated self-attention mechanism into TPP models following modern LLM practices~\citep{zuo2020transformer,zhang2020self,mei2021transformer}, they rely on TPP-specific encoding strategies, including temporal positional encoding for event times and randomly initialized embeddings for event types. This raises the first question: \textbf{how can we coherently integrate standard LLM architectures with TPPs?} We address this by modeling event types and descriptions as textual information and introducing a novel temporal encoding approach using specialized byte-tokens for event time intervals as shown in~\cref{fig:temporal_tokenization}. By converting continuous time intervals to discrete byte-tokens, we can utilize a text template to combine all event information for prompting the LLM. We adapt Qwen2.5~\citep{yang2024qwen25}, a widely used open-source LLM, for this framework. This approach enables straightforward encoding and decoding of event times, types, and descriptions through a language tokenizer.

Recent work, LAMP~\citep{shi2024language}, has demonstrated performance improvements through LLM-based event information reasoning, highlighting the value of textual description information. More recently, TPP-LLM~\citep{liu2024tpp} has shown that integrating LLMs with TPPs can improve prediction accuracy by leveraging textual event descriptions. This leads to the second question: \textbf{what are the benefits of unifying temporal and textual modalities in a unified framework?} We investigate this through two categories of tasks: conventional TPP tasks and LLM-oriented tasks, specifically event description generation. For TPP tasks --- including next event time prediction and type prediction --- we demonstrate state-of-the-art performance through extensive experiments on real-world datasets, comparing against strong baseline models. Notably, on a dataset containing textual event description, our unified framework outperforms LAMP. Furthermore, we find that augmenting temporal information improves event description generation compared to a fine-tuned LLM without temporal information, indicating the benefits of incorporating temporal dynamics into LLMs. 

In summary, our key contributions are as follows:

(1) We introduce a multi-modal framework, \textbf{\BLTPP{}}, that unifies TPPs and LLMs, enabling downstream tasks including event time prediction, type prediction and description generation.

(2) We propose a novel temporal encoding approach using specialized byte-tokens for event time, providing seamless integration with LLM tokenizers.

(3) Through extensive experiments on real-world Web datasets (including AmazonReview, Twitter, StackOverflow, and Taobao), we demonstrate state-of-the-art performance across multiple tasks, with particular emphasis on event description generation—enabling better understanding of Web content and user interactions, a previously unexplored aspect in TPP research.

\section{Related Work}
\label{sec:related_work}

\textbf{Deep TPPs.} Recent advances in TPP modeling have mainly been driven by deep learning-based methods~\citep{bae2023meta,kong2023interval,du2016recurrent,mei2017neural,Mishra2018ModelingPopularity,shchur2019intensity,mei2021transformer}. 
Deep learning-based TPP models have seen significant evolution since~\citep{du2016recurrent}, which first introduced recurrent neural networks (RNNs) for TPP modeling. Early approaches primarily relied on such recurrent architectures, with notable improvements including long short-term memory (LSTM) based models~\citep{mei2017neural} and dual-LSTM frameworks~\citep{xiao2017modeling}. Since the rising popularity of the Transformer architectures~\citep{vaswani2017attention}, attention-based TPPs~\citep{zuo2020transformer,zhang2020self} were introduced, resulting in better modeling of long-range dependencies. Subsequent improvements include non-linear attention score~\citep{zhu2021deep}, sparse attention mechanism~\citep{li2023sparse}, and enhanced interpretability through alignment with statistical nonlinear Hawkes processes~\citep{meng2024interpretable}. 

\textbf{LLM-augmented TPPs.} The integration of LLMs with TPPs represents an emerging frontier in TPP domain. While this direction is still in its early stages, several works demonstrated promising results. \citet{xue2023prompt} introduced PromptTPP, which leverages continual learning principles to enable TPPs to efficiently adapt to streaming event sequences. The fusion of LLMs with TPPs was further advanced by \citet{shi2024language}, which developed a framework that harnesses LLMs' abductive reasoning capabilities to enhance future event prediction. Concurrent to our work, \citet{liu2024tpp} also explored combining LLMs with TPPs through incorporating textual event description and temporal information into pre-trained LLMs. While we share similar high-level goals, our work differentiates itself through a novel temporal encoding strategy using specialized byte-tokens and extends the capability of TPP models to textual event description generation, a previously unexplored direction in TPPs literature. 

\begin{figure*}[!tbp]
    \centering
    \includegraphics[width=0.75\textwidth,page=2]{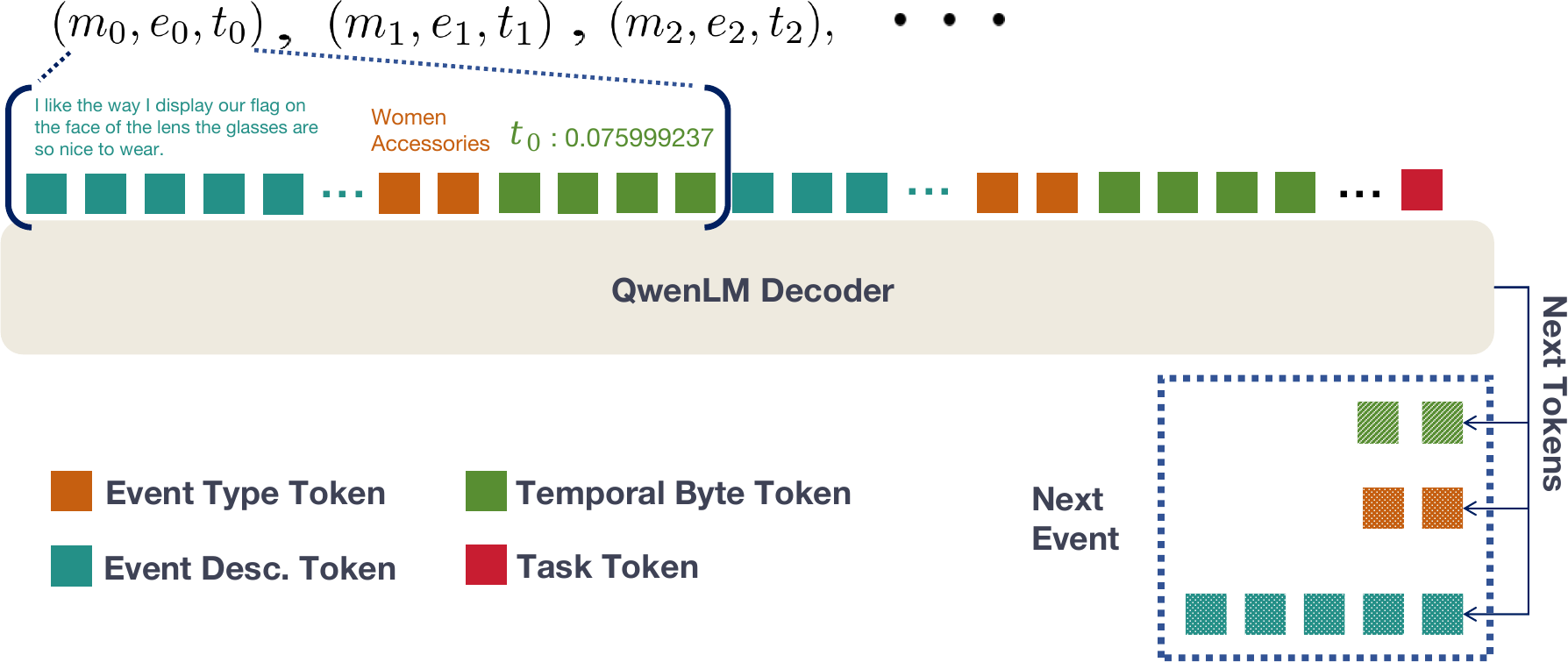}
    \caption{\BLTPP{} processes tokenized event information including event type, description, and time through the QwenLM decoder. The decoder autoregressively generates information about the next event through next-token prediction, while the event intensity is modeled using the hidden state corresponding to the last token.}

    \label{fig:architecture}
    \vspace{-0.5cm}
\end{figure*}

\section{Preliminaries}

This section introduces key concepts and background knowledge for the proposed model, including LLMs and TPPs.
\subsection{Autoregressive Language Model}
State-of-the-art LLMs, such as GPT-4~\citep{achiam2023gpt}, Qwen2.5~\citep{yang2024qwen25} and Gemini 1.5~\citep{team2024gemini}, are built on causal Transformer architectures that process and generate content as discrete tokens. While these models are designed to process textual data via tokenization, handling continuous multi-modal inputs~\citep{hurst2024gpt,yao2024minicpm,fu2025vita}, e.g., images, audio, video, presents unique challenges. To address this, modern architectures typically employ specialized encoders that discretize continuous signals into tokens~\citep{liu2024visual,wang2024qwen2}. For generation in these continuous domains, diffusion models may be used to decode predicted token embeddings back to their original modal representations~\citep{bao2023one,tang2024any}. This encoder-decoder paradigm has become instrumental in bridging the gap between discrete token-based processing and continuous multi-modal data. In this work, we focus on unifying TPPs modality with LLMs, where the primary challenges lie in processing and generating continuous event timestamps.

\subsection{Temporal Point Processes}
TPPs~\citep{Daley2008} are widely used to model the occurrence of events in continuous time. A marked TPP typically associates each event with both an event timestamp $t$ and an event type $e$.
Mathematically, a realization of a marked TPP is a sequence of events ${(t_i,e_i)}_{i=1}^N$ over an observation window $[0,T]$, where $N$ is random and $e_i$ belongs to a discrete set of event types $\mathcal{E}={1,2,\cdots,E}$.
There are two equivalent approaches to modeling TPPs: through the conditional intensity function and through the conditional density function. 
The conditional intensity function is written as:
\begin{equation*}
    \lambda^*(t,e) = \underset{\delta_t \rightarrow 0}{\lim}\frac{p(\text{event with type $e$ in }[t,t+\delta_t)\mid \mathcal{H}_{t}^-)}{\delta_t}.
\end{equation*}
The conditional intensity function specifies the expected number of events occurring within an infinitesimal interval $[t,t+\delta_t)$, given the past history $\mathcal{H}_{t}^-$ up to but not including $t$. 
$*$ indicates the conditioning on history. 
To fit a TPP model to observed data, the log-likelihood function is: 
\begin{equation}~\label{eq:loglikelihood}
    \mathcal{L} = \sum^N_{i=1}\log \lambda^*(t_i,e_i)-\int^T_0 \sum_{e\in \mathcal{E}} \lambda^*(t, e)dt. 
\end{equation}
Alternatively, TPPs can be modeled via the conditional joint density of the next event’s timestamp and type, given the event history:
$
p(t_i, e_i \mid \mathcal{H}_{t_{i-1}})$,
where $\mathcal{H}_{t_{i-1}}$ denotes the history of past events up to time $t_{i-1}$, specifically $(t_1, e_1), \dots, (t_{i-1}, e_{i-1})$. 
In this work, we distinguish between $\mathcal{H}_{t^-}$ and $\mathcal{H}_{t}$: the former refers to the history strictly before time $t$, while the latter includes whether an event occurs exactly at time $t$. 
Using this approach, the log-likelihood function becomes: 
\begin{equation}\label{eq:prob_likelihood}
\mathcal{L} = \sum_{i=1}^{N} \log p(t_i, e_i \mid \mathcal{H}_{t_{i-1}}) + \log \left(1-P(T | \mathcal{H}_{t_{N}}) \right), 
\end{equation}
where $P(T \mid \mathcal{H}_{t_N}) = \int_{t_N}^T \sum_{e=1}^E p(t, e \mid \mathcal{H}_{t_N}) \, \mathrm{d}t$, and the term\((1 - P(T \mid \mathcal{H}_{t_N}))\) represents the probability that no event occurs in \((t_N, T)\). 
As shown by \citet{rasmussen2018lecture}, these two approaches are mathematically equivalent. The conditional density function can be expressed in terms of the conditional intensity function: 
\begin{equation*}
p(t_i, e_i \mid \mathcal{H}_{t_{i-1}}) = \lambda^*(t_i,e_i) \exp\left( - \int_{t_{i-1}}^{t_i} \sum_{e\in \mathcal{E}} \lambda^*(s, e) ds \right)
\end{equation*}
By substituting this expression into~\cref{eq:prob_likelihood}, we obtain the intensity-based likelihood in~\cref{eq:loglikelihood}. 

To capture complex event dynamics, neural networks have been introduced to parametrize $\lambda^*(t,e)$~\citep{mei2017neural} or $p(t, e \mid \mathcal{H}_{t_{i}})$~\citep{shchur2019intensity}. This allows for the direct learning of temporal dependencies and event type distributions from the data. In deep TPPs, an embedding layer is introduced to map each event $(t_i, e_i)$ to a dense vector $\mathbf{z}_i \in \mathbb{R}^D$. This embedding encodes both temporal and event type information, serving as the foundation for modeling sequential dependencies.
By employing an RNN or Transformer, we can model the dependency by summarizing the embeddings of observed events into a history representation $(\mathbf{h}_i \in \mathbb{R}^M)$ in a recurrent or autoregressive manner. This history embedding $\mathbf{h}_i$ can then be used either to model the conditional intensity function $\lambda^*(t,e)$ or the conditional density function $p(t, e \mid \mathbf{h}_i)$ for predicting the next event time and type. 

In our work, we leverage the conditional density function approach as it naturally aligns with the generative capabilities of LLMs through next-token predictions. In particular, by discretizing the continuous time distribution through byte-tokenization, we enable LLMs to model and sample from $p(t, e \mid \mathbf{h}_i)$ directly.

\section{Language Modeling with TPP}
In this section, we first present \BLTPP{}, which bridges TPPs and natural language, along with the corresponding data preprocessing procedure. We then detail the inference and training methodology. 

\subsection{Modeling}
We consider a more general event sequence dataset that includes descriptive information about events, represented as \(\{(t_i, e_i, m_i)\}_{i=1}^N\), where \(t_i\) denotes the event timestamp, \(e_i\) represents the event type, and \(m_i\) is the textual description of the event. 
Our goal is to model the dependencies in such an event sequence and predict the times, types, and descriptions of future events.

In this work, we adopt a sequence-to-sequence model. Specifically, as depicted in~\cref{fig:architecture}, the model backbone is a causal decoder-only Transformer based on Qwen2.5~\citep{yang2024qwen2}. 
We convert each event $(t_i, e_i, m_i)$ into a sequence of tokens. 
In simple terms, for the event type \(e_i\) and description \(m_i\), we employ a built-in language tokenizer; for the event timestamp \(t_i\), we design a specialized byte-level tokenization strategy. 
We explain the tokenization in detail as follows. 

\textbf{Event description tokenization.}
The event description $m_i$ is in natural language, so we can directly use a built-in language tokenizer for tokenization. 

\textbf{Event type tokenization.}
If the event type in the dataset is represented in natural language, we can directly use a built-in tokenizer.  
If the event type is represented as an index, we attempt to recover its text label from the index. For example, \(e_i = 0\) represents \textit{Women Shoes} in the \textit{Amazon Review} dataset~\citep{ni2019justifying}.  
If the event type is represented as an index and no corresponding text label is provided, we directly use the index as the text label. 

\begin{figure}[t]
    \centering
    \includegraphics[width=0.49\textwidth,page=3]{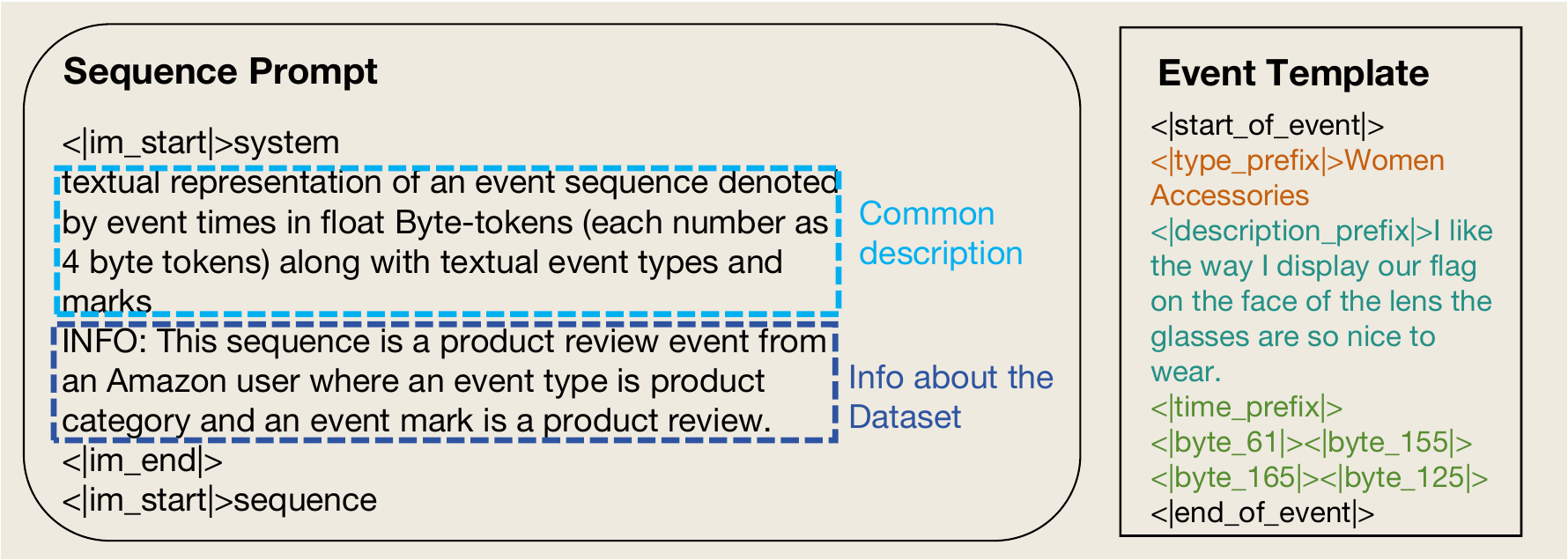}
    \caption{Illustration of the textual template used to convert an event sequence into the input for the language model. Prepended by the sequence prompt, the event template structures each event with its type, description, and timestamp.}
    \label{fig:template}
    \vspace{-0.5cm}
\end{figure}

\textbf{Event timestamp tokenization.}
For event timestamps, a naive approach is to directly treat the timestamp \(t_i\) as text. However, this method requires a large number of tokens to represent a single timestamp, significantly reducing the model's processing efficiency.  
Inspired by prior works in byte-to-byte models~\citep{xue2022byt5,pagnoni2024byte}, we propose a specialized byte-level tokenization strategy. 
Specifically, we augment the vocabulary of the Qwen2.5 model with $256$ new special byte tokens, ``$<|byte\_x|>$'' $\forall x \in \{0,1,\cdots,255\}$, representing all unique values of a single byte. 
We can then parse a 32-bit floating point precision number (timestamp) into $4$ byte tokens\footnote{In implementation, we do not directly perform byte tokenization on timestamps but instead on event intervals, which facilitates subsequent time prediction tasks.}. 
This is more token-efficient compared to directly tokenizing the timestamp as text. 
For example, as shown in~\cref{fig:temporal_tokenization}, it takes $11$ tokens to represent ``0.075999237'' using the default Qwen2.5 tokenizer whereas it only takes $4$ byte tokens using our approach, i.e., ``$<|byte\_125|><|byte\_165|><|byte\_155|><|byte\_61|>$''. 

\textbf{Event template.}
Eventually, we concatenate all encoded components within a predefined textual template as demonstrated in~\cref{fig:template} where each event is surrounded by special tokens. Specifically, we use ``$<|start\_of\_event|>$'' and ``$<|end\_of\_event|>$'' to denote the start and end of an event, respectively. We use ``$<|description\_prefix|>$'', ``$<|type\_prefix|>$'' and ``$<|time\_prefix|>$'' to prepend the event description, type and temporal byte tokens. 
Please refer to~\cref{sec:prompt}
for details about the templates and samples of generated prompts from datasets. 

\subsection{Inference}
\BLTPP{} can be applied to various downstream tasks, including event time prediction, type prediction, and description generation.
We first define a set of task tokens, ``$<|description\_prediction|>$'', ``$<|type\_prediction|>$'' and ``$<|time\_prediction|>$''.
When conducting a specific downstream task, we concatenate the corresponding task token to the end of the token sequence. 
We then autoregressively predict the next tokens until the end-of-sequence token is reached. 
We apply a task-specific decoding strategy to obtain the desired output. 

\textbf{Event time prediction}: We obtain the predicted next event interval $\tau_{i+1}$ by decoding the temporal byte tokens back to a float pointing number. We then obtain the next event time, $t_{i+1} = t_i + \tau_{i+1}$. 

\textbf{Event type prediction} and \textbf{event description generation}: 
    We obtain the desired textual information by decoding the generated tokens back into natural language, as done in LLMs.

\subsection{Training}
We collect several widely adopted public TPP datasets for training the model. We follow the standard training precedure of GPT-like LLMs~\citep{openai2022}, and design our training protocol as two stages. 

\textbf{Stage 1: Continued pre-training.} We use the fixed template in~\cref{fig:template} to convert the event sequence to the token sequence $(x_1,x_2,\ldots,x_L)$.  
At this stage, we conduct a continued pre-training of the LLM backbone on the token sequence with the next-token prediction loss: 
$
    \mathcal{L}_{\text{stage1}}(\theta) = - \frac{1}{L} \sum_{l=1}^{L} \log p_\theta(x_l | x_{<l})$.

\textbf{Stage 2: Next-event fine-tuning.} We augment the model with capabilities to conduct downstream tasks on TPPs. 
We generate training samples by randomly sampling a segment of the event sequence as the prompt \((p_1, p_2, \ldots)\), with the corresponding next event as the response \((r_1, \ldots, r_R)\). This approach constructs prompt-response pairs for event time prediction, type prediction, and description generation. 
We compute the next-token prediction loss on the response: 
$\mathcal{L}_{\text{stage2}}(\theta) = - \frac{1}{R} \sum_{l=1}^{R} \log p_\theta(r_l | p_1,p_2,\ldots, r_{<l})$ at training.

It is worth noting that, unlike common TPP training approaches, which involve training on a specific dataset and then testing on the test set of the same dataset, we merge all datasets into a single combined dataset for training and then evaluate the model separately on each individual test set.

\section{Experiments}
To evaluate the effectiveness of \BLTPP{}, we conduct extensive experiments comparing against several baselines on real-world TPP datasets. 
We also examine the impact of different model components and training protocol through an ablation study.

\subsection{Baselines}
We employ several previous works as baselines for comparison.
\begin{itemize}
\item \textbf{Neural Hawkes Process (NHP)}~\citep{mei2017neural}:
NHP employs a continuous-time LSTM to encode the temporal and type information of historical events. The resulting history embedding is used to model the conditional intensity function. This model cannot handle or generate event descriptions. 
\item \textbf{Self-Attentive Hawkes Process (SAHP)}~\citep{zhang2020self}, \textbf{Transformer Hawkes Process (THP) and Attentive Neural Hawkes Process (AttNHP)}~\citep{zuo2020transformer}:
SAHP, THP and AttNHP all encode historical events through self-attention mechanism, and uses the history embedding to model the conditional intensity function. They cannot handle or generate event descriptions.
\item \textbf{Intensity-free TPP (IFTPP)}~\citep{shchur2019intensity}: IFTPP directly modeling the conditional
distribution of inter-event intervals similar to our work. 
This model cannot handle event descriptions or generate event descriptions.
\item \textbf{TPP-LLM (TinyLlama-1.1B)}~\citep{liu2024tpp}: TPP-LLM integrates LLMs with TPPs by directly utilizing textual event type descriptions and incorporating temporal embeddings with parameter-efficient fine-tuning. While it can process textual descriptions for prediction tasks, it cannot generate future event descriptions.

\item \textbf{GPT-3.5-turbo enhanced Attentive Neural Hawkes Process (ANHP-G3.5)}~\citep{shi2024language}: the best performing baseline in their proposed LAMP, a method that leverages pre-trained LLMs for abductive reasoning to improve event prediction. 
This model can handle historical event descriptions but cannot generate future event descriptions. 
\item \textbf{Qwen2.5-0.5B}~\citep{yang2024qwen2}:
As far as we know, no existing TPPs baseline can generate future event descriptions.
To evaluate the quality of event description generation, we compare our method with the vanilla language model Qwen2.5-0.5B, fine-tuned on the same dataset without temporal information. 
\end{itemize}

\begin{table}[t]
\centering
\caption{Dataset statistics. Train/Dev/Test shows the number of sequences in each split, $K$ is the number of event types, $L_{avg}$ represents the average sequence length.}
\label{tab:dataset_stats}
\begin{tabular}{lcccc}
\toprule
Dataset & Train/Dev/Test & $K$ & $L_{avg}$ & Event Desc. \\
\midrule
\textit{Retweet} & 9000 / 1535 / 1520 & 3 & 70 & \ding{55} \\
\textit{Stackoverflow} & 1400 / 400 / 400 & 22 & 65 & \ding{55} \\
\textit{Taobao} & 1300 / 200 / 500 & 17 & 150 & \ding{55} \\
\textit{Taxi} & 1400 / 200 / 400 & 10 & 37 & \ding{55} \\
\textit{Amazon Review} & 2434 / 1633 / 1952 & 24 & 27 & $\checkmark$ \\
\bottomrule
\end{tabular}
\end{table}

\begin{table*}[t]
\centering
\caption{Prediction performance comparison on real-world datasets for event times and types. Results reported in terms of RMSE and ACC with standard deviations. Best results are in bold.}
\label{tab:main_results}
\begin{tabular}{lccccccc}
\toprule
\multirow{2}{*}{Dataset} & \multicolumn{6}{c}{RMSE$(\downarrow)$ / ACC$(\uparrow)$} \\
\cmidrule{2-8}
& NHP & SAHP & THP & IFTPP & AttNHP & TPP-LLM & \BLTPP{}-0.5B  \\
\midrule
\multirow{2}{*}{\textit{Retweet}} & 21.8 / 54.0 & 21.7 / 54.0 & 25.3 / 58.5  & 22.2 / \textbf{60.3} & 22.2 / 59.9 & 21.3 / 54.1 &  \textbf{18.1} / 59.7 \\
& 0.184 / 0.002 & 0.301 / 0.002 & 0.188 / 0.003 & 0.180/0.003 & 0.204 / 0.003 & 0.002 / 0.002 & 0.002 / 0.001  \\
\midrule
\multirow{2}{*}{\textit{StackOverflow}} & 1.37 / 45.0  & 1.38 / 43.9 & 1.37 / 45.0  &  1.37 / 44.9 & 1.37 / 44.8 & 1.17 / 41.9 & \textbf{1.12} / \textbf{45.5}\\
& 0.011 / 0.006 & 0.013 / 0.005 & 0.021 / 0.006 &  0.010 / 0.005 & 0.019/0.003 & 0.002 / 0.000 & 0.001 / 0.002 \\
\midrule
\multirow{2}{*}{\textit{Taobao}} & 0.55 / 57.0 & 0.53 / 50.3  & 0.34 / 57.9  & 0.53 / 56.6 & 0.30 / 46.3  & 0.32 / 43.57 & \textbf{0.21} / \textbf{59.7} \\
& 0.005 / 0.006 & 0.004 / 0.002 & 0.003 / 0.004 & 0.005 / 0.004 & 0.005/0.001 & 0.001 / 0.002 & 0.002 / 0.002 \\
\midrule
\multirow{2}{*}{\textit{Taxi}} & 0.37 / \textbf{91.5} &  0.37 / 90.3 & 0.37 / 91.3  & 0.38 / 91.4  & 0.37 / 91.3 & 0.43 / 91.2 & \textbf{0.32} / 90.5 \\
& 0.003 / 0.0003 & 0.003 / 0.0005 & 0.003 / 0.0008  & 0.003 / 0.006 & 0.003/0.000 & 0.001 / 0.000 & 0.002 / 0.000  \\
\bottomrule
\end{tabular}
\end{table*}

\subsection{Datasets}
\label{ssec:datasets}

We use five real-world datasets to evaluate the performance of our method and the baselines.
\begin{itemize}
    
\item  \textbf{Retweet}~\citep{zhou2013learning}: A dataset of time-stamped user retweet event sequences. Events are categorized into three types based on user follower counts: \textit{small} ($< 120$ followers), \textit{medium} ($120-1363$ followers), and \textit{large} ($> 1363$ followers).
\item \textbf{Stackoverflow}~\citep{jure2014snap}: Two years of user award records from Stackoverflow, where each sequence represents events of granting badges to a user. There are in total $22$ different badges. 
\item \textbf{Taobao}~\citep{xue2022hypro}: Time-stamped user click behaviors on Taobao e-commerce platform, collected over a nine-day period. Events represent item clicks, with types corresponding to item categories.
\item \textbf{Taxi}~\citep{whong2014foiling}: Time-stamped taxi pick-up and drop-off events across the five boroughs of New York City. Each combination (borough, pick-up or drop-off) defines an event type, resulting in 10 distinct event types. 
\item \textbf{Amazon Review}~\citep{ni2019justifying}: A dataset of product reviews from Amazon, containing sequences from the 2,500 most active users. Events represent product reviews, with types corresponding to product categories. The $23$ most frequent categories were preserved as distinct types, with remaining categories merged into one. 
Each event includes a text description containing the review. 
\end{itemize}

We note that among the five datasets, the first four are traditional TPP datasets that include only event times and types, while \textit{Amazon Review} is the only dataset containing textual event descriptions. Therefore, we use this dataset for event description evaluations.
The splitting and statistics of all datasets are provided in \cref{tab:dataset_stats}.

\subsection{Metrics}
We employ the following metrics across different TPP tasks:
\begin{itemize}
\item For event time prediction, we use root-mean squared error (\textbf{RMSE}) to measure the temporal prediction accuracy, i.e., $\sqrt{\sum_{i=1}^N(\tau_i - \hat{\tau_i})^2/N}$, $\hat{\tau}_i$ is the predicted time interval.
\item For event type prediction, we report the prediction accuracy (\textbf{ACC}), i.e., $\sum_{i=1}^N \mathbbm{1}_{e_i=\hat{e}_i}/N$, where $\hat{e}_i$ is the predicted event type.
\item We evaluate the quality of generated event descriptions as a standard NLP task, employing \textbf{ROUGE-L} scores~\citep{lin2004rouge}. These scores assess the overlap between model-generated and ground-truth text based on $n$-grams, providing a comprehensive measure of both precision and recall. 
\item We analyze the sentiment of generated descriptions using \textbf{VADER} scores~\citep{hutto2014vader}, which provide compound polarity scores ranging from $-1$ (most negative) to $+1$ (most positive) in sentiment. 
\end{itemize}

\subsection{Experimental Setup}
Our proposed model is built on top of Qwen2.5-0.5B~\citep{yang2024qwen25}.
We utilize the base model variant rather than its instruction-tuned version, as our implementation employs a different prompting template from the standard chat template. We use a single NVIDIA A40 for model training and experiments. At each stage, we train the model for $5$ epochs with learning rate $1e^{-4}$ and select checkpoints achieving best validation losses. Details about hyperparameters used in training can be found in~\cref{sssec:hyper}. We sample $50$k prompt-response pairs for next-event fine-tuning (stage 2).
The baseline models are implemented using the open source frameworks, including EasyTPP~\citep{xue2023easytpp} and LAMP~\citep{shi2024language} with default configurations. We reuse partial experimental results from~\citep{xue2023easytpp,shi2024language}. 
\begin{figure}[t]
    \centering
    \begin{minipage}[b]{0.23\textwidth}
        \centering
        \subfloat[]{\label{fig:rouge_a}
        \includegraphics[width=0.9\textwidth,page=1]{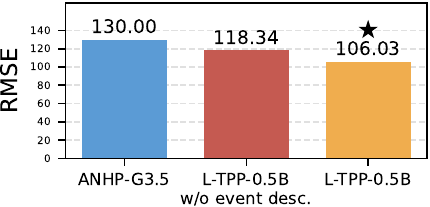}}\\[0.5em]
        \subfloat[]{\label{fig:rouge_b}
        \includegraphics[width=0.9\textwidth,page=2]{figures/figure_ab_pages.pdf}}
    \end{minipage}
    \hfill
    \begin{minipage}[b]{0.22\textwidth}
        \centering
        \subfloat[]{\label{fig:rouge_c}
        \includegraphics[width=\textwidth]{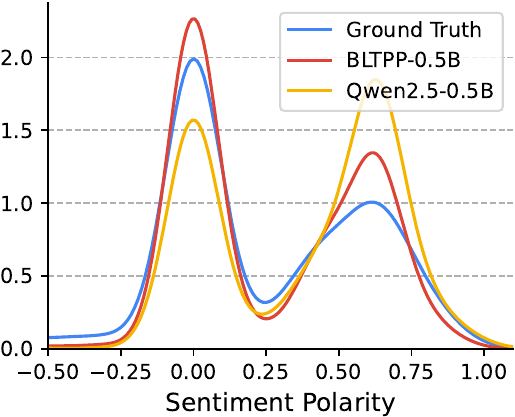}}
    \end{minipage}
    \caption{Results on Amazon Review dataset: (a) The RMSE$(\downarrow)$ on event time prediction; (b) the ROUGE-L$(\uparrow)$ score on textual event description generation; (c) comparison of sentiment polarity distribution of generated event descriptions.}
    \label{fig:rouge}
    \vspace{-0.2cm}
\end{figure}

\subsection{Results and Analysis}

\textbf{Performance on type-marked TPP:} 
In this experiment, we study the conventional TPP tasks on type-marked TPPs in terms of event time prediction and type prediction.
\cref{tab:main_results} presents the results for event prediction tasks, including time and type predictions. \BLTPP{}-0.5B demonstrates superior or competitive performance on both RMSE and ACC metrics, with notably low standard deviations across all datasets.

For event time prediction, \BLTPP{}-0.5B shows substantial improvements on RMSE across most datasets. On \textit{Retweet}, \BLTPP{}-0.5B achieves $18.1$, significantly outperforming all baselines including the recent TPP-LLM ($21.3$) and AttNHP ($22.2$), as well as traditional methods (ranging from $21.7$ to $25.3$). Similar improvements are observed on \textit{Stackoverflow} ($1.12$ versus $1.17$--$1.38$) and \textit{Taobao} ($0.21$ versus $0.30$--$0.55$), where \BLTPP{}-0.5B substantially outperforms both the LLM-based baseline TPP-LLM and attention-based methods. On \textit{Taxi}, \BLTPP{}-0.5B achieves the best RMSE of $0.32$, outperforming TPP-LLM ($0.43$) and matching or exceeding traditional TPP methods.
For event type prediction, \BLTPP{}-0.5B achieves the highest ACC on \textit{Stackoverflow} ($45.5\%$) and \textit{Taobao} ($59.7\%$), surpassing both TPP-LLM ($41.9\%$ and $43.57\%$ respectively) and AttNHP ($44.8\%$ and $46.3\%$ respectively). On \textit{Retweet}, \BLTPP{}-0.5B ($59.7\%$) shows competitive performance compared to IFTPP's best result ($60.3\%$) and AttNHP ($59.9\%$), while substantially outperforming TPP-LLM ($54.1\%$). On the \textit{Taxi} dataset, \BLTPP{}-0.5B ($90.5\%$) performs slightly below NHP's ($91.5\%$) but remains competitive with other baselines including AttNHP ($91.3\%$) and TPP-LLM ($91.2\%$).

These results demonstrate that \BLTPP{}-0.5B provides robust improvements in temporal prediction while maintaining or enhancing event type prediction capabilities across diverse real-world datasets. Notably, \BLTPP{}-0.5B consistently outperforms TPP-LLM, another LLM-based approach, highlighting the effectiveness of our byte-token temporal encoding mechanism. We note that the consistently low standard deviations of \BLTPP{}-0.5B results are due to the low temperature (0.0) during inference.

\begin{table}[t]
    \centering
    \caption{Ablation study of the impact of tokenization approaches, training strategies, and LLM sizes. Results reported in terms of RMSE and ACC.} 
    \label{tab:ablation-training}
    \resizebox{\columnwidth}{!}{%
    \begin{tabular}{ccccc}
    \toprule
    \multirow{2}{*}{Models} & \multicolumn{4}{c}{RMSE$(\downarrow)$ / ACC$(\uparrow)$} \\
    \cmidrule{2-5}
    & \textit{Retweet} & \textit{Stackoverflow} & \textit{Taobao} & \textit{Taxi} \\
    \midrule
    \BLTPP{}-1.5B (Qwen2.5) & 18.3 / 56.4 & \textbf{1.10} / 44.1 & 0.26 / 58.2 & 0.33 / 90.4 \\
    \BLTPP{}-1B (Gemma-3) & 18.8 / 58.6	& 1.17 / 41.2 &	\textbf{0.20} / 58.8	 & \textbf{0.31} / 90.4\\
    \BLTPP{}-0.5B (Qwen2.5) & \textbf{18.1} / \textbf{59.7} & 1.12 / \textbf{45.5} & 0.21 / \textbf{59.7} & 0.32 / \textbf{90.5} \\
    w/o Stage 1 training & 19.2 / 58.5 & 1.20 / 44.3 & 0.28 / 57.8 & 0.34 / 89.8 \\
    w/o byte tokens & 21.8 / 57.4 & 1.34 / 44.1 & 0.35 / 59.4 & 0.34 / 90.4 \\
    \bottomrule
    \end{tabular}
    }
    \end{table}

\textbf{Performance on description-marked TPP:} 
In this part, we design a new type of experiment for description-marked TPPs. 
We use \BLTPP{}-0.5B to process the time, type, and description of historical events, and attempt to predict the time, type, and description of future events. 
We present results on the \textit{Amazon Review} dataset in~\cref{fig:rouge} where each event is marked with a product review. Specifically, we use the summary of the review as the textual description in the experiments. 
We first investigate the prediction performance in~\cref{fig:rouge_a} where we compare \BLTPP{}-0.5B against the recent LLM-augmented model ANHP-G3.5~\citep{shi2024language}. The results show that \BLTPP{}-0.5B performs significantly better than the baseline model, achieving an RMSE of $106.03$ compared to $130.00$ for ANHP-G3.5. 
In particular, we also present RMSE for \BLTPP{}-0.5B trained without event description, which achieves $118.34$, demonstrating that incorporating event description enhances the model's temporal prediction capability. 

In~\cref{fig:rouge_b}, we study a novel event description generation task, where we evaluate \BLTPP{}-0.5B's ability to generate review summaries using the ROUGE-L metric. 
\BLTPP{}-0.5B achieves a ROUGE-L score of $24.78$, surpassing the baseline Qwen2.5-0.5B that is fine-tuned on the same dataset without temporal information, which scores $22.60$. 
This improvement suggests that jointly learning temporal dynamics and textual information leads to better quality in generated review summaries.

To further assess the quality of generated event descriptions, we provide both concrete examples and systematic sentiment analysis. For instance, for a product in the ``Children Accessories'' category, \BLTPP{}-0.5B generates contextually appropriate descriptions such as ``Perfect for my 3 year old'' which aligns with both the product category and typical review patterns.

While individual examples demonstrate the model's capability to generate reasonable descriptions, we conduct a more comprehensive evaluation through sentiment polarity analysis using a rule-based sentiment analyzer~\citep{hutto2014vader}, as shown in \cref{fig:rouge_c}. The results exhibit a consistent bimodal distribution across all three models, with dominant peaks at neutral and positive sentiments. Notably, Qwen2.5-0.5B underestimates neutral sentiment while overestimating positive sentiment. In contrast, \BLTPP{}-0.5B yields a distribution that more closely aligns with the ground truth, indicating a better preservation of the natural sentiment patterns inherent in the review data---likely due to its modeling of temporal dynamics.

\begin{figure}[t]
    \centering
    \subfloat[]{\label{fig:event_dis_a}
    \includegraphics[width=0.49\columnwidth,page=1]{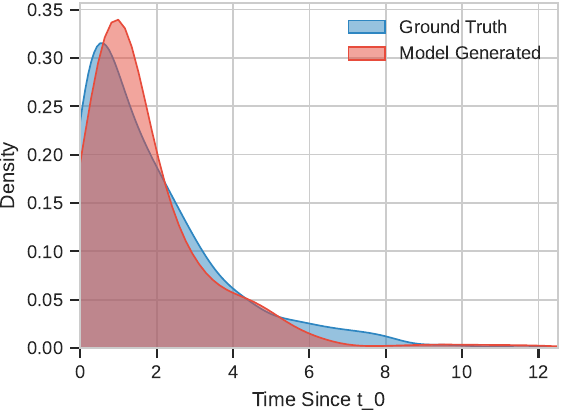}}
    \hfill
    \subfloat[]{\label{fig:event_dis_b}
    \includegraphics[width=0.49\columnwidth,page=2]{figures/event_distributions.pdf}}
    \caption{Comparison of ground-truth and model-generated distributions for StackOverflow with $e_0=3$ and $t_0=0$: (a) Dist. of event interval $\tau_1$; (b) Dist. of event type $e_1$.} 
    \label{fig:event_dis}
\end{figure}

\subsection{Ablation Studies}
\label{ssec:ablation}
In this section, we present an ablation study on \BLTPP{} by removing or altering key components and settings, as shown in~\cref{tab:ablation-training}. 
We analyze the impact of tokenization approaches, training strategies, and LLM sizes. 

\textbf{Tokenization approaches:} 
We first study the contribution of byte-tokens to model performance. 
In this experiment, we replace temporal byte tokens with standard tokenization, where time intervals are rounded to three decimal places and represented as strings. 
All other configurations, including prompting templates and hyperparameters, remain constant. When using standard string tokenization, we observe consistent performance degradation across all datasets. Most notably, on the \textit{Retweet} dataset, RMSE increased significantly from $18.1$ to $21.8$, while ACC dropped by $2.26\%$. This demonstrates that byte-token representation enables more precise temporal pattern learning compared to standard tokenization.

\textbf{Training strategies:} 
Given that our goal is to predict future events based on observed event sequences --- a task directly addressed in next-event fine-tuning (stage 2) --- a natural question arises: is stage 1 pre-training necessary? To answer this, we conduct an ablation study where we maintain identical model architectures and hyperparameters while varying the training procedure. The base model (w/o Stage 1 training) is directly fine-tuned on downstream tasks without continued pre-training on temporal sequences.
Comparing this to the full model with all training stages, we observe consistent improvements across all datasets. The addition of Stage 1 training leads to substantial improvements, particularly in the \textit{Retweet} dataset (RMSE decrease from $19.2$ to $18.1$) and \textit{Taobao} dataset (RMSE decrease from $0.28$ to $0.21$). These results validate the importance of our staged training approach in enhancing the model's temporal understanding. 

\textbf{LLM variants and sizes:} We experiment with two model sizes (0.5B and 1.5B) with Qwen2.5~\citep{yang2024qwen25} and one for Gemma-3 from Google~\citep{team2025gemma} (1B).
Interestingly, the smaller models often achieve better performance than their larger counterparts across different datasets. The Qwen2.5-0.5B model shows strong performance on \textit{Retweet} (RMSE: $18.1$, ACC: $59.7\%$) and \textit{Stackoverflow} (ACC: $45.5\%$), while the Gemma-3-1B model outperforms others on \textit{Taobao} (RMSE: $0.20$) and \textit{Taxi} (RMSE: $0.31$). The Qwen2.5-1.5B model performs best only on \textit{Stackoverflow} RMSE ($1.10$).
This counter-intuitive result might be attributed to the mismatch between model capacity and dataset size. While larger language models perform well at general language tasks, the relatively limited size of TPP datasets may not provide sufficient supervision for effectively adapting larger models to this specialized temporal modeling task. We hypothesize that smaller models might achieve better parameter efficiency in learning task-specific patterns from limited data while maintaining good generalization capability.

These ablation studies demonstrate that our design choices, particularly the byte-token representation and staged training strategy, are crucial for the model's performance.
We also show that relatively small model architectures are sufficiently effective for TPP modeling tasks.

\subsection{Qualitative Analysis}

To further evaluate \BLTPP{}'s capability in capturing underlying data distributions from Web event sequences, we examine the conditional distributions of event intervals and types. This analysis is particularly relevant for understanding how well the model captures user behavior on Web platforms, where temporal dynamics and event type distributions reflect real user interactions.

Specifically, we visualize the conditional distribution of the second event—given a fixed first event at $t_0 = 0$—for both ground-truth data and model-generated samples, as shown in~\cref{fig:event_dis_a} and~\cref{fig:event_dis_b}. We focus on the second event because all sequences in the dataset share the same event history at this point, i.e., $(t_0, e_0)$. As a result, the second event across sequences follows the same conditional distribution, making it suitable for reliable comparison. In contrast, later events are conditioned on diverse historical contexts, resulting in heterogeneous conditional distributions that are less amenable to aggregated analysis.

Using the \textit{Stackoverflow} dataset and conditioning on $e_0 = 3$ (corresponding to a specific question category), we generate 2,000 samples with \BLTPP{}-0.5B. The resulting conditional distributions show excellent alignment with those of the ground truth across both temporal and categorical dimensions:

\textbf{Temporal distribution (\cref{fig:event_dis_a}):} The distribution of inter-event times exhibits a characteristic right-skewed pattern common in Web user activity, with most events occurring within a short time window after the initial event. \BLTPP{} accurately captures this temporal pattern, including the peak density around 1-2 time units and the long tail extending to 12+ time units. This demonstrates the model's ability to learn realistic temporal dynamics of user engagement on Q\&A platforms, where follow-up activities (e.g., answers, comments) typically occur soon after a question is posted, with occasional delayed responses.

\textbf{Event type distribution (\cref{fig:event_dis_b}):} The categorical distribution shows strong alignment between generated and ground-truth samples across all event types. The model correctly identifies the dominant event type (type 4, with probability $\sim$0.6) and accurately reproduces the relative frequencies of less common event types. Notably, \BLTPP{} captures the multi-modal nature of the distribution, with secondary peaks at types 9 and 0, reflecting the diverse types of user interactions that follow an initial question post on \textit{Stackoverflow}.

The close match between model-generated and ground-truth distributions underscores \BLTPP{}'s effectiveness in modeling both the temporal dynamics and mark distributions of real-world Web event sequences. This capability is crucial for applications such as user behavior prediction, content recommendation timing, and understanding engagement patterns on Web platforms. Unlike traditional TPP models that may struggle with complex, multi-modal distributions, our LLM-based approach leverages the representational power of large language models to capture nuanced patterns in Web user behavior.

\subsection{Scalability to Long Event Sequences}

A practical consideration for deploying \BLTPP{} on Web platforms is its ability to handle long event sequences, which are common in real-world scenarios such as long-term customer purchase histories. With Qwen2.5-0.5B's context window of 32k tokens, our method can handle sequences of up to 3,000 events in a single forward pass. For longer sequences, standard techniques like sliding windows can be employed. A key advantage of our byte-token encoding strategy is its compatibility with existing LLM inference optimization frameworks (e.g., vLLM~\citep{kwon2023efficient}, FlashAttention~\citep{dao2022flashattention}), allowing us to leverage state-of-the-art acceleration techniques without customization.

To evaluate potential performance degradation on longer sequences, we employ perplexity, a standard metric for assessing long-context capabilities of LLMs~\citep{chen2023extending,chen2023clex}, to measure sequence modeling performance across different token lengths. \cref{tab:perplexity} presents the perplexity scores on the \textit{Amazon Review} test set, stratified by content length.
The results demonstrate a gradual performance degradation as sequence length increases, with perplexity rising from 5.74 for short sequences (0-300 tokens) to 9.36 for longer sequences (900-1200 tokens). This trend is consistent with the behavior of standard LLMs on long-context tasks~\citep{liuetal2023}. Despite this degradation, the perplexity remains at reasonable levels even for sequences approaching 1,200 tokens, indicating that \BLTPP{} maintains effective modeling capabilities. %

\begin{table}[t]
\centering
\caption{Perplexity (PPL) of \BLTPP{} on \textit{Amazon Review} sequences with varying content lengths. Lower perplexity indicates better performance.}
\label{tab:perplexity}
\begin{tabular}{lcccc}
\toprule
\textbf{No. tokens} & \textbf{0-300} & \textbf{300-600} & \textbf{600-900} & \textbf{900-1200} \\
\midrule
PPL & 5.74 & 7.08 & 8.20 & 9.36 \\
\bottomrule
\end{tabular}
\end{table}

\section{Conclusions}

In this paper, we presented \BLTPP{}, a unified framework that bridges TPPs with LLMs for modeling event sequences with both temporal and textual information on Web platforms. By introducing specialized byte-tokens for temporal encoding and leveraging text templates for event representation, our framework enables seamless integration of TPPs with standard LLM architectures. Extensive experiments on real-world Web datasets—including e-commerce reviews, social media posts, and online community interactions—demonstrate that \BLTPP{} achieves state-of-the-art performance on conventional TPP tasks while enabling high-quality event description generation, a capability previously unexplored in TPP literature. Our results highlight the mutual benefits of combining temporal dynamics with LLMs, offering new opportunities for understanding and analyzing temporal patterns in Web-scale data.
The implications of this work extend to various Web applications, including personalized recommendation systems, content moderation, and user behavior analysis. By jointly modeling temporal dynamics and textual content, \BLTPP{} provides a more comprehensive understanding of how events unfold on Web platforms, which can inform the design of more responsive and context-aware Web services.

\textbf{Limitations and future work:} 
One limitation of the proposed model is the potential context length explosion when processing events with lengthy textual descriptions. Additionally, the current framework may face scalability issues when handling extremely large-scale event sequences with other modalities, such as images and audio. Future work could address these limitations through more sophisticated encoding strategies and attention mechanisms, while also exploring more complex multi-modal information and investigating the framework's applicability to larger datasets.
In particular, the development of large-scale multi-modal datasets for TPP research remains an important future direction.

\begin{acks}
This work was supported by the NSFC Project (No.62576346), the MOE Project of Key Research Institute of Humanities and Social Sciences (22JJD110001), the fundamental research funds for the central universities, and the research funds of Renmin University of China (24XNKJ13), and Beijing Advanced Innovation Center for Future Blockchain and Privacy Computing. 
\end{acks}

\bibliographystyle{ACM-Reference-Format}
\bibliography{references}

\appendix

\section{Prompt Details}\label{sec:prompt}

We present the detailed prompt templates used for preprocessing temporal point process datasets. The templates consist of system prompts and event-specific formats. 

\textbf{System prompt templates:} there are two variants of system templates used in the preprocessing where the original number prompt is used in~\cref{ssec:ablation}. 

\begin{table}[h]
\small
\begin{tabular}{|p{0.95\columnwidth}|}
\hline
\textbf{Byte-token Prompt:} \\
$<|im\_start|>$system\\
textual representation of an event sequence denoted by event times in float Byte-tokens (each number as 4 byte tokens) along with textual event types\\
INFO: \{sequence\_info\}\\
$<|im\_end|>$\\
$<|im\_start|>$sequence \\
\hline
\textbf{Original Number Prompt:} \\
$<|im\_start|>$system\\
textual representation of an event sequence denoted by event times in float numbers along with textual event types\\
INFO: \{sequence\_info\}\\
$<|im\_end|>$\\
$<|im\_start|>$sequence \\
\hline
\end{tabular}
\end{table}
where \{sequence\_info\} refers to dataset-specific information. 

\begin{table}[h]
\small
\begin{tabular}{|l|p{0.7\columnwidth}|}
\hline
\textbf{Dataset} & \textbf{Information} \\
\hline
\textit{StackOverflow} & This sequence is a sequence of badges awarded to a user in StackOverflow. There are 22 event types. \\
\hline
\textit{Retweet} & This sequence is a sequence of retweets of a tweet. There are 3 event types. \\
\hline
\textit{Taobao} & This sequence is a sequence of clicks on a product in Taobao. There are 20 event types. \\
\hline
\textit{Taxi} & This sequence tracks taxi pick-up and drop-off events. There are 10 event types. \\
\hline
\textit{Amazon Review} & This sequence is a product review event from an Amazon user where event type is product category. \\
\hline
\end{tabular}
\label{tab:dataset_info}
\end{table}
After the system prompt,
individual events are formatted using the following template:

\begin{table}[h]
\small
\begin{tabular}{|p{\columnwidth}|}
\hline
$<|start\_of\_event|>$\\
$<|type\_prefix|>$\{event\_type\}\\
$<|description\_prefix|>$\{event\_description\}\\
$<|time\_prefix|>$\{event\_time\}\\
$<|end\_of\_event|>$ \\
\hline
\end{tabular}
\end{table}
where \{event\_type\}, \{event\_description\} and \{event\_time\} are textual content of an event in a temporal point process. 

\textbf{Special tokens:} we present all special tokens added to the Qwen2.5 tokenizer vocabulary and used in our work to structure the prompts in~\cref{tab:special_tokens}. 

\begin{table}[htp]
\small
\centering
\caption{Special tokens used in prompt templates.}
\label{tab:special_tokens}
\begin{tabular}{|l|p{0.49\columnwidth}|}
\hline
\textbf{Special Token} & \textbf{Description} \\
\hline
$<|start\_of\_event|>$ & Tokens for marking the start of an event \\
$<|end\_of\_event|>$ & Tokens for marking the end of an event \\
\hline
$<|type\_prefix|>$ & Prefix token for event type \\
$<|description\_prefix|>$ & Prefix token for event description \\
$<|time\_prefix|>$ & Prefix token for event timestamp \\
\hline
$<|type\_prediction|>$ & Task token for type inference \\
$<|description\_prediction|>$ & Task token for description inference \\
$<|time\_prediction|>$ & Task token for time inference \\
\hline
$<|byte\_0|>$ to $<|byte\_255|>$ & Byte tokens for representing event time intervals as float32 numbers \\
\hline
\end{tabular}
\end{table}

\textbf{Event sequence samples:} we provide two samples of generated event sequences from \textit{Amazon Review} dataset in~\cref{tab:event_sample} and \textit{StackOverflow} in~\cref{tab:event_sample_2}, respectively. These are used in training stage 1. We also show a sample of prompt-response pair from \textit{Amazon Review} used in next-event fine-tuning in~\cref{tab:prompt_response}. We note that $<|im\_start|>$ and $<|im\_end|>$ are built-in special tokens from the Qwen2.5 tokenizer. 

\begin{table}[h]
\small
\caption{Event sequence sample from \textit{Amazon Review}.}
\label{tab:event_sample}
\begin{tabular}{|p{0.95\columnwidth}|}
\hline
\vspace{0.1cm}
\textbf{Prompt:} \\
$<|im\_start|>$system\\
textual representation of an event sequence denoted by event times in float Byte-tokens (each number as 4 byte tokens) along with textual event types\\
INFO: This sequence is a product review event from an Amazon user where event type is product category\\
$<|im\_end|>$\\
$<|im\_start|>$sequence\\
$<|start\_of\_event|>$\\
$<|type\_prefix|>$Luggage \& Travel Gear\\
$<|description\_prefix|>$Great bag..Scooby in pink!\\
$<|time\_prefix|>$$<|byte\_0|>$$<|byte\_0|>$$<|byte\_0|>$$<|byte\_0|>$\\
$<|end\_of\_event|>$\\
$<|start\_of\_event|>$\\
$<|type\_prefix|>$Children Shoes\\
$<|description\_prefix|>$Twinkle toes is the best!\\
$<|time\_prefix|>$$<|byte\_68|>$$<|byte\_42|>$$<|byte\_0|>$$<|byte\_0|>$\\
$<|end\_of\_event|>$\\
$<|start\_of\_event|>$\\
$<|type\_prefix|>$Women Jewelry\\
$<|description\_prefix|>$Pretty earrings.\\
$<|time\_prefix|>$$<|byte\_66|>$$<|byte\_140|>$$<|byte\_0|>$$<|byte\_0|>$\\
$<|end\_of\_event|>$\\
$<|start\_of\_event|>$\\
$<|type\_prefix|>$Men Uniforms, Work \& Safety\\
$<|description\_prefix|>$Nice, work shirt.\\
$<|time\_prefix|>$$<|byte\_66|>$$<|byte\_156|>$$<|byte\_0|>$$<|byte\_0|>$\\
$<|end\_of\_event|>$\\
\hline
\end{tabular}
\end{table}

\begin{table}[h]
\caption{Event sequence sample from \textit{StackOverflow}.}
\label{tab:event_sample_2}
\small
\begin{tabular}{|p{0.95\columnwidth}|}
\hline
\vspace{0.1cm}
\textbf{Prompt:} \\
$<|im\_start|>$system \\
textual representation of an event sequence denoted by event times in float Byte-tokens (each number as 4 byte tokens) along with textual event types \\
INFO: This sequence is a sequence of badges awarded to a user in \textit{StackOverflow}. There are 22 event types. \\
$<|im\_end|>$ \\
$<|im\_start|>$sequence \\
$<|start\_of\_event|>$ \\
$<|type\_prefix|>$5 \\
$<|time\_prefix|>$$<|byte\_0|>$$<|byte\_0|>$$<|byte\_0|>$$<|byte\_0|>$ \\
$<|end\_of\_event|>$ \\
$<|start\_of\_event|>$ \\
$<|type\_prefix|>$13 \\
$<|time\_prefix|>$$<|byte\_58|>$$<|byte\_240|>$$<|byte\_0|>$$<|byte\_0|>$ \\
$<|end\_of\_event|>$ \\
$<|start\_of\_event|>$ \\
$<|type\_prefix|>$3 \\
$<|time\_prefix|>$$<|byte\_64|>$$<|byte\_134|>$$<|byte\_225|>$$<|byte\_0|>$ \\
$<|end\_of\_event|>$ \\
$<|start\_of\_event|>$ \\
$<|type\_prefix|>$3 \\
$<|time\_prefix|>$$<|byte\_62|>$$<|byte\_222|>$$<|byte\_48|>$$<|byte\_0|>$ \\
$<|end\_of\_event|>$ \\
$<|start\_of\_event|>$ \\
$<|type\_prefix|>$2 \\
$<|time\_prefix|>$$<|byte\_61|>$$<|byte\_82|>$$<|byte\_128|>$$<|byte\_0|>$ \\
$<|end\_of\_event|>$ \\
$<|start\_of\_event|>$ \\
$<|type\_prefix|>$3 \\
$<|time\_prefix|>$$<|byte\_64|>$$<|byte\_109|>$$<|byte\_212|>$$<|byte\_0|>$ \\
$<|end\_of\_event|>$ \\
$<|start\_of\_event|>$ \\
$<|type\_prefix|>$3 \\
$<|time\_prefix|>$$<|byte\_62|>$$<|byte\_142|>$$<|byte\_64|>$$<|byte\_0|>$ \\
$<|end\_of\_event|>$ \\
$<|start\_of\_event|>$ \\
$<|type\_prefix|>$5 \\
$<|time\_prefix|>$$<|byte\_63|>$$<|byte\_50|>$$<|byte\_24|>$$<|byte\_0|>$ \\
$<|end\_of\_event|>$ \\
\hline
\end{tabular}
\end{table}

\begin{table}[h]
\caption{Prompt-response pair sample from \textit{Amazon Review}.}
\label{tab:prompt_response}
\small
\begin{tabular}{|p{0.95\columnwidth}|}
\hline
\vspace{0.1cm}
\textbf{Prompt:} \\
$<|im\_start|>$system \\
textual representation of an event sequence denoted by event times in float Byte-tokens (each number as 4 byte tokens) along with textual event types \\
INFO: This sequence is a product review event from an Amazon user where event type is product category \\
$<|im\_end|>$ \\
$<|im\_start|>$sequence \\
$<|start\_of\_event|>$$<|type\_prefix|>$Men Surf, Skate \& Street$<|description\_prefix|>$I am a long-time fan of Reef sandals \\
$<|time\_prefix|>$$<|byte\_0|>$$<|byte\_0|>$$<|byte\_0|>$$<|byte\_0|>$ \\
$<|end\_of\_event|>$ \\
$<|start\_of\_event|>$$<|type\_prefix|>$Men Shoes$<|description\_prefix|>$I own 8 pair of Allen Edmonds and I like them all.  They are very comfortable \\
$<|time\_prefix|>$$<|byte\_67|>$$<|byte\_18|>$$<|byte\_0|>$$<|byte\_0|>$ \\
$<|end\_of\_event|>$ \\
$<|start\_of\_event|>$$<|type\_prefix|>$Shoe, Jewelry \& Watch Accessories$<|description\_prefix|>$Easy to use \\
$<|time\_prefix|>$$<|byte\_65|>$$<|byte\_144|>$$<|byte\_0|>$$<|byte\_0|>$ \\
$<|end\_of\_event|>$ \\
$<|start\_of\_event|>$$<|type\_prefix|>$Men Shoes$<|description\_prefix|>$Great Shoe \\
$<|time\_prefix|>$$<|byte\_65|>$$<|byte\_224|>$$<|byte\_0|>$$<|byte\_0|>$ \\
$<|end\_of\_event|>$ \\
$<|start\_of\_event|>$ \\
$<|time\_prediction|>$ \\
\hline
\vspace{0.1cm}
\textbf{Response:} \\
$<|byte\_61|>$$<|byte\_130|>$$<|byte\_13|>$$<|byte\_139|>$ \\
\hline
\end{tabular}
\end{table}

\section{Datasets}\label{ap:datasets}
We provide extra details about the splitting and statistics of the datasets here:
\begin{itemize}
    \item For the conventional TPP datasets, including \textit{Retweet}, \textit{Stackoverflow}, \textit{Taobao} and \textit{Taxi}, we obtain the processed datasets from \url{http://bit.ly/40seop9} following the splitting setups in~\citep{xue2023easytpp}. 
    \item \textit{Amazon Review}: The data is chronologically split into training (before 2015-08-01), validation (2015-08-01 to 2016-02-01), and test (after 2016-02-01) sets following setups from~\citet{shi2024language}. 
\end{itemize}

\section{Experimental Setups}
\label{sssec:hyper}

\textbf{Hyperparameters:} we use similar hyperparameter settings for each training stage shown in~\cref{tab:hyperparams}. All stages use mixed-precision training with bfloat16, evaluate and save checkpoints at the end of each epoch, and keep only the best performing model. Stage 2 specifically uses accuracy as the metric for selecting the best model.
\begin{table}[H]
\centering
\caption{Training hyperparameters for different stages.}
\label{tab:hyperparams}
\begin{tabular}{lccc}
\toprule
\textbf{Hyperparameter} & \textbf{Stage 1} & \textbf{Stage 2} & \textbf{Stage 3} \\
\midrule
Learning Rate & 1e-4 & 1e-4 & 1e-4 \\
Batch Size & 4 & 4 & 32 \\
Weight Decay & 0.01 & 0.01 & 0.01 \\
Number of Epochs & 5 & 5 & 5 \\
Gradient Accumulation Steps & 4 & 4 & 4 \\
\bottomrule
\end{tabular}
\end{table}

\newpage

\end{document}